%% file: neurips_2024.tex
\documentclass{article}

\usepackage[dblblindworkshop, final]{neurips_2025}

\title{Foundation Models as World Models: A Foundational Study in Text-Based GridWorlds}
\workshoptitle{Embodied World Models for Decision Making}

\author{
  Remo Sasso \quad Michelangelo Conserva \quad Dominik Jeurissen \quad Paulo Rauber \\
  School of Electronic Engineering and Computer Science\\
  Queen Mary University of London, United Kingdom \\
  \texttt{\{r.sasso, m.conserva, d.jeurissen, p.rauber\}@qmul.ac.uk}
}

\usepackage[utf8]{inputenc} % allow utf-8 input
\usepackage[T1]{fontenc}    % use 8-bit T1 fonts
\usepackage{hyperref}       % hyperlinks
\usepackage{url}            % simple URL typesetting
\usepackage{booktabs}       % professional-quality tables
\usepackage{amsfonts}       % blackboard math symbols
\usepackage{nicefrac}       % compact symbols for 1/2, etc.
\usepackage{microtype}      % microtypography
\usepackage{xcolor}         % colors
\usepackage{graphicx}
\usepackage{amsmath}
\usepackage{amssymb}
\usepackage{mathtools}
\usepackage{amsthm}
\usepackage{wrapfig}
\usepackage{listings}
\usepackage{svg}
\usepackage{caption}
\usepackage{subcaption}
\usepackage{algorithmic}
\usepackage{algorithm}
\usepackage[capitalize,noabbrev]{cleveref} % Load after hyperref

\definecolor{darkblue}{rgb}{0, 0, 0.5}
\hypersetup{colorlinks=true, citecolor=darkblue, linkcolor=darkblue, urlcolor=darkblue}
\definecolor{codegray}{rgb}{0.5,0.5,0.5}
\definecolor{codepink}{RGB}{252, 142, 172}
\definecolor{codepurple}{rgb}{0.58,0,0.82}
\definecolor{backcolour}{RGB}{245,245,245}
\lstdefinestyle{mystyle}{
    backgroundcolor=\color{backcolour},   
    commentstyle=\color{magenta},
    keywordstyle=\color{blue},
    numberstyle=\tiny\color{codegray},
    stringstyle=\color{codepurple},
    basicstyle=\fontfamily{\ttdefault}\footnotesize,
    breakatwhitespace=false,         
    breaklines=true,                 
    keepspaces=true,    
    frame=single,
    numbersep=5pt,                  
    showspaces=false,                
    showstringspaces=false,
    showtabs=false,                  
    tabsize=2,
    classoffset=1, % starting new class
    keywordstyle=\color{violet},
    classoffset=0,
}
\lstdefinelanguage{JavaScript}{
  keywords={typeof, new, true, false, catch, function, return, null, catch, switch, var, if, in, while, do, else, case, break},
  keywordstyle=\color{black},
  ndkeywords={class, export, boolean, throw, implements, import, this},
  ndkeywordstyle=\color{darkgray}\bfseries,
  identifierstyle=\color{black},
  sensitive=false,
  comment=[l]{//},
  morecomment=[s]{/*}{*/},
  commentstyle=\color{purple}\ttfamily
}

\begin{document}

\maketitle

\input{sections/0_abstract}
\input{sections/1_introduction}
\input{sections/2_related_works}
\input{sections/4_methods}
\input{sections/5_experiments}
\input{sections/6_conclusion}

\newpage

\onecolumn
\bibliography{neurips_2024}
\bibliographystyle{icml2023}

\newpage

\input{sections/8_appendix}

\end{document}

%% file: sections/0_abstract.tex
\begin{abstract}
    While reinforcement learning from scratch has shown impressive results in solving sequential decision-making tasks with efficient simulators, real-world applications with expensive interactions require more sample-efficient agents.
    Foundation models (FMs) are natural candidates to improve sample efficiency as they possess broad knowledge and reasoning capabilities, but it is yet unclear how to effectively integrate them into the reinforcement learning framework.
    In this paper, we anticipate and, most importantly, evaluate two promising strategies.
    First, we consider the use of foundation world models (FWMs) that exploit the prior knowledge of FMs to enable training and evaluating agents with simulated interactions.
    Second, we consider the use of foundation agents (FAs) that exploit the reasoning capabilities of FMs for decision-making.
    We evaluate both approaches empirically in a family of grid-world environments that are suitable for the current generation of large language models (LLMs).
    Our results suggest that improvements in LLMs already translate into better FWMs and FAs; that FAs based on current LLMs can already provide excellent policies for sufficiently simple environments; and that the coupling of FWMs and reinforcement learning agents is highly promising for more complex settings with partial observability and stochastic elements.
\end{abstract}

%% file: sections/1_introduction.tex
\section{Introduction} \label{sec:introduction}
While reinforcement learning agents have achieved impressive results in mastering complex games \citep{berner2019dota, vinyals2019grandmaster, Schrittwieser2019-df}, their applicability is often limited by sample inefficiency, requiring millions of interactions to learn effective policies. A promising solution is to incorporate prior knowledge, removing the constraint of learning \textit{tabula rasa}. Foundation models (FMs), with their vast pre-trained knowledge and reasoning capabilities, are natural candidates for this role \citep{brown2020language, team2023gemini, achiam2023gpt}.

However, it is unclear how to best integrate these powerful models. Prior work has often used them as auxiliary components for reward shaping \citep{klissarov2024motif} or as high-level planners controlling a set of abstract skills \citep{llm_survey}. This paper investigates a more direct form of integration by asking a fundamental question: to what extent can contemporary FMs serve as zero-shot world models or low-level actors? We evaluate two competing strategies:
\begin{itemize}
    \item \textbf{Foundation World Models (FWMs):} Using an FM as a zero-shot simulator to generate interaction data for pre-training a traditional RL agent.
    \item \textbf{Foundation Agents (FAs):} Using an FM directly as a decision-making policy, generating low-level actions at each step.
\end{itemize}

To isolate the core simulation and reasoning capabilities of FMs, we conduct our analysis in a controlled "laboratory": a family of text-based grid-world environments. This deliberate choice allows us to systematically probe their abilities in handling deterministic and stochastic dynamics without the confounding complexities of visual perception. Our work contributes a direct, comparative analysis of the FWM and FA approaches at the level of low-level dynamics simulation and action selection. By deliberately avoiding intermediate abstractions like planners or predefined skills, we provide a foundational benchmark of the zero-shot capabilities of the foundation models themselves.

Our results reveal a clear trade-off. In simple, deterministic settings, FAs provide excellent zero-shot policies. However, for more complex, stochastic, and partially observable tasks, the FWM-RL approach is far more robust, demonstrating that pre-training on simulated data provides a substantial sample efficiency gain. Our findings provide a foundational analysis for the burgeoning field of world models, offering key insights into how these powerful priors can be harnessed to build more efficient agents.

%% file: sections/2_related_works.tex
\section{Related Work} \label{sec:related_works}

\paragraph{Foundation models for simulation.}
Prior knowledge and generative capabilities can allow foundation models to generate environment dynamics for a broad range of applications.
\citet{uni_sim} developed a high-quality interactive physics simulator trained from video data only.
\citet{generative_agents} showed that LLMs can simulate the behavior of autonomous agents deployed in a small town.
Another line of work explores LLM simulations in combination with planning algorithms.
\citet{guan2023leveraging} and \citet{valmeekam2023on} demonstrated that LLMs can produce simulations with a planning domain definition language (PDDL) that can be combined with standard solvers.
Alternatively, LLMs can directly act as forward models for tree search methods
\citep{hao2023reasoning, hao2024llm, yao2024tree}.
While these results show that LLM simulations can be successfully used for planning, their applicability is often limited by domain compatibility with PDDL and the need for perfect simulations.  

Recent work has systematically benchmarked the simulation ability of LLMs. \citet{wang2024bytesized32} introduced \textit{ByteSized32}, showing that while GPT-4 can handle simple text-based transitions, it is unreliable on arithmetic and implicit-physics dynamics. To improve robustness, neurosymbolic approaches have emerged: \citet{tang2024worldcoder} propose \textit{WorldCoder}, where LLMs iteratively write Python code to represent environment dynamics, and \citet{dainese2024gifmcts} combine LLM-generated code with MCTS refinement to yield verified executable world models. At the multimodal scale, DeepMind’s \textit{Genie 3} \citep{deepmind2025genie3} demonstrates that foundation world models can now generate interactive 3D environments at 24fps with persistent object memory and emergent physics, representing a new frontier for simulation-based pretraining.

\paragraph{Foundation models for decision-making.}
Beyond simulation, LLMs can act directly as agents. \citet{liu2023rafa} proposed \textit{Reason for Future, Act for Now (RAFA)}, where an LLM imagines possible futures while a policy executes near-term actions, improving robustness over naive direct control. Other approaches augment decision-making with structured search, such as Tree-of-Thoughts \citep{yao2023tree} and Graph-of-Thoughts \citep{besta2024graph}, which extend chain-of-thought prompting into trees or graphs of possible action sequences. These methods highlight that while foundation agents excel in short, deterministic tasks, more systematic reasoning is needed for long-horizon or stochastic domains. Multimodal foundation models further expand this scope by enabling agents to act in environments with both text and images \citep{driess2023palm, lin2023text2motion}.

\paragraph{Foundation models as auxiliary components.}
In reinforcement learning, foundation models can also guide agents indirectly through reward shaping and goal generation. \citet{klissarov2024motif} introduced an intrinsic reward model based on LLM preferences, substantially improving exploration in sparse-reward environments. \citet{eureka} demonstrated that if environment code is available, an LLM can program and refine a reward function based on agent feedback. \citet{yuqing2023guidingpretraining} used an LLM to generate plausible goals for a goal-conditioned agent, while \citet{emma} trained a multimodal RL agent to imitate an expert LLM policy defined via PDDL. More recently, \citet{ma2024fme} proposed \textit{Foundation Model Exploration}, where a vision-language model intermittently proposes exploratory actions, yielding substantial gains in sparse-reward domains with minimal prompting.  

In our work, we specifically focus on the direct integration of LLMs in reinforcement learning via low-level action selection and environment dynamics simulation.

%% file: sections/4_methods.tex
\section{Methods} \label{sec:methods}

In this section, we introduce the methodology proposed to evaluate the direct integration of LLMs within the reinforcement learning framework.
Section \ref{sec:test_envs} describes the family of grid world environments used for evaluation.
Section \ref{sec:fwm} describes how we induce LLMs to simulate these grid worlds to facilitate a foundation world model and delineates the challenges involved.
Finally, Section \ref{sec:fa} integrates LLMs in the decision-making process of reinforcement learning through two strategies: combining the LLM-based world model with reinforcement learning agents and directly querying LLMs for low-level action selection.

\subsection{Grid World} \label{sec:test_envs}
In this paper, we specifically focus on grid world environments, which is one of the most commonly used environment types in reinforcement learning \citep{MinigridMiniworld23}.
In addition to the simplicity of encoding them in textual representation, grid worlds have the advantage of being sufficiently simple to allow in-depth investigations and can be adapted to include substantial difficulties. Furthermore, grid worlds inherently pose interesting challenges for LLMs, as simulation and planning require the handling of numerical operations and constraints such as grid boundaries.

We define a grid world as a square grid of size \(n \times n\), where \(n\) is a positive integer.
A state is defined as a tuple of two integers $[x,y]$ encoding the coordinate position of the agent inside the grid.
The agent's starting position is fixed at the bottom-left corner. 
The maximum number of time steps is set to $2n^2$. 
The agent has a set of four possible actions to move within the grid: \(\mathcal{A} = \{\text{up, down, left, right}\}\).
When the agent takes an action that attempts to go outside of the boundaries, it has no effect.
There exists a single reward location that yields a reward of 1, all other locations yield a reward of 0. 

We consider two grid world instances designed to highlight the strengths and weaknesses of LLMs for simulation and decision-making.
The first one is deterministic and fully observable.
The reward location is fixed in the top-right corner, and this information is included in the input for the decision-making agents.
This allows us to examine the ability of LLMs to simulate the environment dynamics and how well they can utilize their reasoning abilities to locate and navigate to the reward.
The second setting is stochastic and partially observable.
The reward location is randomly sampled every episode, and its location is not provided to the agents.
This setting is significantly more challenging to solve for both simulation and decision-making.
For simulation, the LLMs are now also required to reproduce a uniform sampling distribution of reward locations.
Additionally, while in the previous setting, a decision-making agent is only required to navigate to the reward location, in this setting they need to systematically explore the grid.  
In Appendix \ref{app:more_elements} we consider further variations of the grid world, such as adding a key or incorporating sticky actions.

\subsection{Foundation World Models} \label{sec:fwm}

\paragraph{Simulating environment dynamics.}
% To explore leveraging the broad knowledge of LLMs to construct a model of the environment, w.
A world model is composed of a transition function, which describes how the environment state changes in response to an action of the agent, and a reward function, which specifies the immediate rewards corresponding to state transitions.
To reproduce these functions with an LLM, we can design a prompt template containing a description of the environment, the transition function, and the reward function.
The variables in the template are the grid size <n>, the location of the reward \textsc{<REWARD LOCATION>}, the current state of the environment \textsc{<OBSERVATION>}, and the agent action \textsc{<ACTION>}.

\paragraph{Prompt template.}
To provide a brief contextualization of the problem, we first describe the grid world environment.
\begin{lstlisting}[breaklines=true, language=JavaScript]
Imagine a gridworld of size <n> by <n> with an observation system of coordinates, where [0,0] is the bottom left corner. 
\end{lstlisting}
For the reward function, we can simply list the two possible outcomes, such that the LLM is tasked with associating the current location of the agent with the reward location.
\begin{lstlisting}[breaklines=true, language=JavaScript]
Determine which reward condition the next observation meets, and respond the corresponding reward and terminal status:
    - The observation is the same as <REWARD LOCATION> - 1 and True
    - The observation is not the same as <REWARD LOCATION> - 0 and False
\end{lstlisting}
The transition function of the grid world is the most challenging component as it is not limited to comparison but also requires the execution of bounded mathematical operations.
For this reason, we consider two prompt templates for the grid world transition function that we refer to as $T_\text{minimal}$ and $T$.
The former only provides a minimalist description of the transition function and more heavily relies on the LLMs' prior knowledge of grid worlds and inference capabilities:
\begin{lstlisting}[breaklines=true, language=JavaScript]
Determine the next observation after taking action <ACTION> from <OBSERVATION> with the following rules:
    - The location [0,0] is the bottom-left corner, and [<n-1>,<n-1>] it the top-right corner.
\end{lstlisting}
The latter includes a more detailed description of the transition function and the mathematical constraints that can be leveraged by the LLMs reasoning capabilities:
\begin{lstlisting}[breaklines=true, language=JavaScript]
Determine the next observation after taking action <ACTION> from <OBSERVATION> with the following rules:
        - The x and y values are always non-negative and lower than <n-1>.
        - 'right' means increment x
        - 'left' means decrement x 
        - 'up' means increment y
        - 'down' means decrement y.
        - Actions have no effect when they go outside the boundaries.
\end{lstlisting}

\paragraph{Temperature.} 
In this paper, we assume $T$ and $R$ to be deterministic. 
This means that even though we use generative models that are inherently non-deterministic, we need to encourage as little diversity in the responses as possible. 
The primary factor that affects the variability in the responses of large language models (LLMs) is the temperature parameter, denoted as $\tau$, which generally ranges from 0 to 2. 
This parameter essentially determines the likelihood that the model will choose the token with the highest probability instead of selecting among other possible tokens.
Thus, we can use $\tau = 0$ to encourage the simulation of deterministic functions $T$ and $R$. 
However, reinforcement learning environments may contain stochastic elements. 
For instance, in the second variation of our grid world the reward location is randomly sampled from a uniform distribution in each episode. 
To simulate this through FWMs is non-trivial, as ensuring that the distribution of outcomes produced by an LLM matches a specified distribution is challenging \citep{papamarkou2024position}. 
Nonetheless, we investigate the ability of LLMs to facilitate this stochastic element by setting $\tau=1.8$ to introduce significantly more variability and uncertainty into the model's responses. 
We then prompt the LLM with a specification of the desired distribution as: 
\begin{lstlisting}[breaklines=true, language=JavaScript]
Generate a random location on an <n> by <n> grid and respond it as a coordinate [x,y].
\end{lstlisting}
We investigate the simulation of stochastic elements with non-uniform distributions in Appendix \ref{app:non-uniform}.

\subsection{Decision-making agents} \label{sec:fa}

\paragraph{FWM-based reinforcement learning.} 
% In model-based reinforcement learning, agents develop an internal model of the environment dynamics, enabling them to predict the outcomes of their actions. 
Model-based reinforcement learning agents aim to learn an explicit model of the environment dynamics.
Utilizing these world models, agents can simulate environment interactions, which often results in significantly better sample efficiency compared to their model-free counterparts \citep{Ha2018-jd, Kaiser2019-ts, janner2019trust, Schrittwieser2019-df, sasso2023posterior, hafner2023mastering}. 
In our approach, we pre-train the agents using simulated interactions from the FWM before letting the agent interact with the real environment.
This integration leverages the prior knowledge of LLMs along with the adaptive decision-making capabilities of reinforcement learning algorithms.
We focus on Policy Gradient (PG) agents \citep{sutton2018reinforcement}, being a popular choice among practitioners.
More specifically, we use Trust Region Policy Optimization (TRPO) \citep{janner2019trust} for the deterministic grid and Proximal Policy Optimization \citep{schulman2017proximal} with a recurrent neural network (RecurrentPPO), for the stochastic and partially observable grid world.

\paragraph{Foundation agents.}
The prior knowledge and reasoning abilities encoded in LLMs provide the potential to induce zero-shot optimal policies.
For example, common sense naturally suggests that the action 'up' induces a unitary increase of the $y$ coordinate and that the grid should be systematically explored to find a random reward location.
Therefore, we investigate how well LLMs can exploit their capabilities to directly act in the environment as such Foundation Agents (FAs).
We do so through three different prompt templates, that include the description of the environment, the objective, and the full history of previously visited locations, but differ in the level of guidance to utilize this information:
\begin{itemize}
    \item \textbf{Action Only (AO)}: The agent is asked to respond with an action given the current state and information available.
    \item \textbf{Simple Plan (SP)}: The agent is encouraged to first reason about what it should do next given the current state and information available, then decide on an action based on that plan. 
    \item \textbf{Focused Plan (FP)}: The agent is explicitly told to use its memory to determine what position it wants to go next, then analyze which action is best to efficiently reach that position, and finally respond with that action.
\end{itemize}
These three prompts aim to encourage the agent to use different levels of reasoning and planning strategies, helping to assess how effectively LLMs can navigate decision-making tasks and adjust their behavior based on varying levels of guidance and complexity.
Particularly in the stochastic setting used in this paper, the agent should utilize its memory to know what locations it previously visited to consistently find the reward.
The AO prompt provides minimal guidance and relies on the model figuring out itself that it should utilize its memory to decide on the next action. 
The SP prompt increases the level of guidance by encouraging the model to first reason with the available information and generate a plan accordingly. 
Finally, the FP prompt explicitly tells the model to generate a plan by utilizing its memory of previously visited states and respond with an action based on that plan. 
The corresponding prompt templates and further implementation details can be found in Appendix \ref{app:prompt_templates_llms}. 

%% file: sections/5_experiments.tex
\section{Experiments} \label{sec:experiments}
In this section, we perform an in-depth examination of the FWM and FA concepts through empirical experiments. 
Section \ref{sec:fwm_experiments} provides an in-depth study of FWMs through simulation experiments and Section \ref{sec:fa_experiments} examines FAs and the combination of FWMs with reinforcement learning agents through decision-making experiments. 

\paragraph{Foundation models.}
The LLMs used for all examinations in this paper include GPT-3.5 and GPT-4 \citep{achiam2023gpt}, Gemma 2b and Gemma 7b \citep{team2024gemma}, and Gemini 1.0 and Gemini 1.5 \citep{team2023gemini}. This diverse collection enables us to assess the simulation and decision-making abilities of LLMs across various model types and generations. Although the exact parameter counts for the GPT and Gemini models are unknown, performance benchmarks suggest that Gemini 1.0 aligns closely with GPT-3.5, and Gemini 1.5 with GPT-4 in terms of parameter range \citep{chiang2024chatbot}. The Gemma models, rooted in the same research and technology as the Gemini models, serve as a reference point for smaller model sizes. 

\begin{figure}[t]
    \centering
    \includegraphics[width=\linewidth]{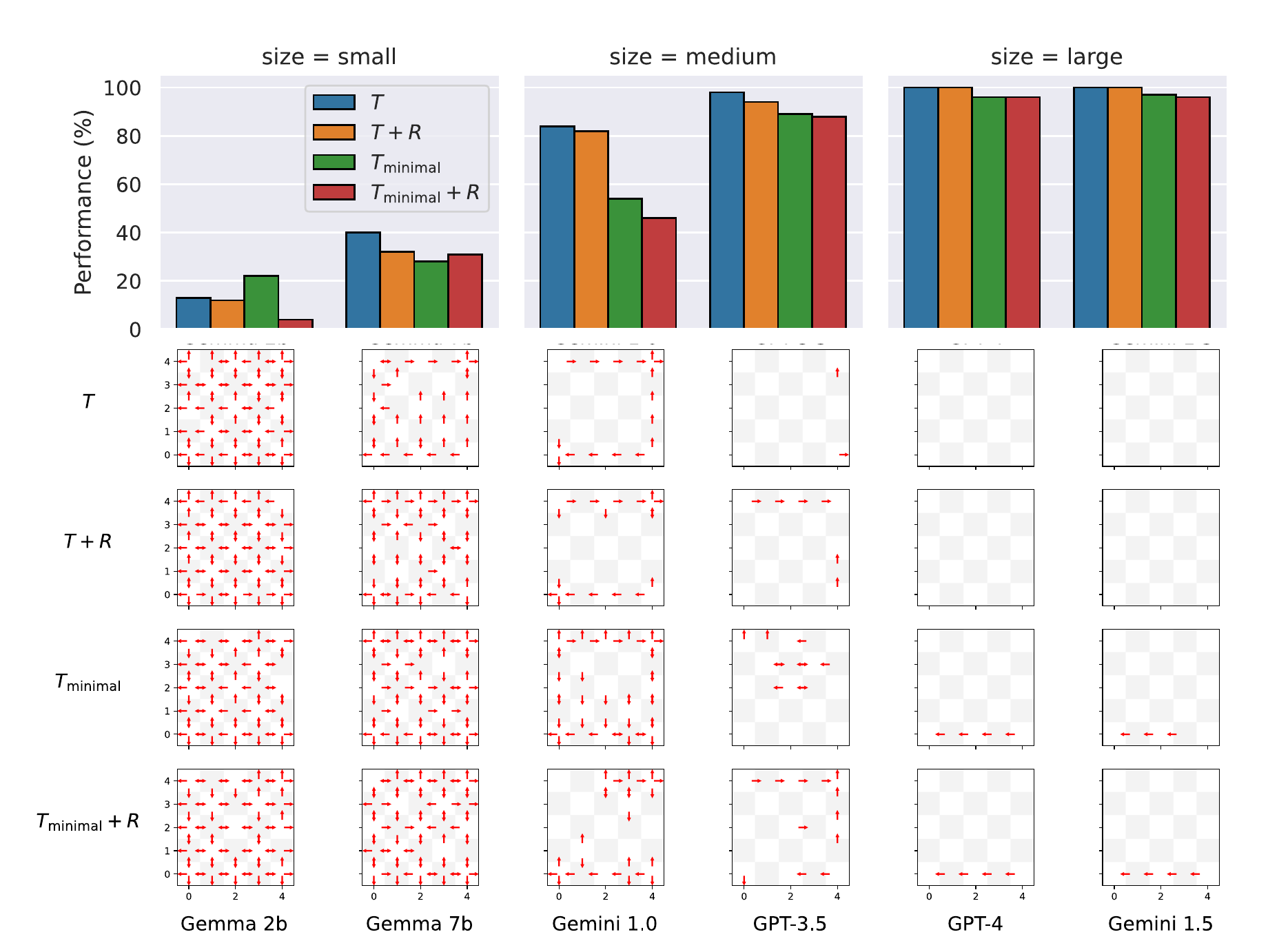}
    \caption{Simulation performance for LLMs with various prompt approaches. The bar plots show the simulation accuracy for each combination. Below each model's bar plot, the corresponding errors are visualized as red arrows in the grid. Prompt $T$ queries the transition function with detailed descriptions, $T + R$ queries both the transition and reward function with detailed descriptions, $T_\text{minimal}$ queries only the transition function with a minimal description, and prompt $T_\text{minimal} + R$ adds the reward function to the latter.}
    \label{fig:sim_combined}
\end{figure}

\begin{figure}[t]
    \centering
    \includegraphics[width=\linewidth]{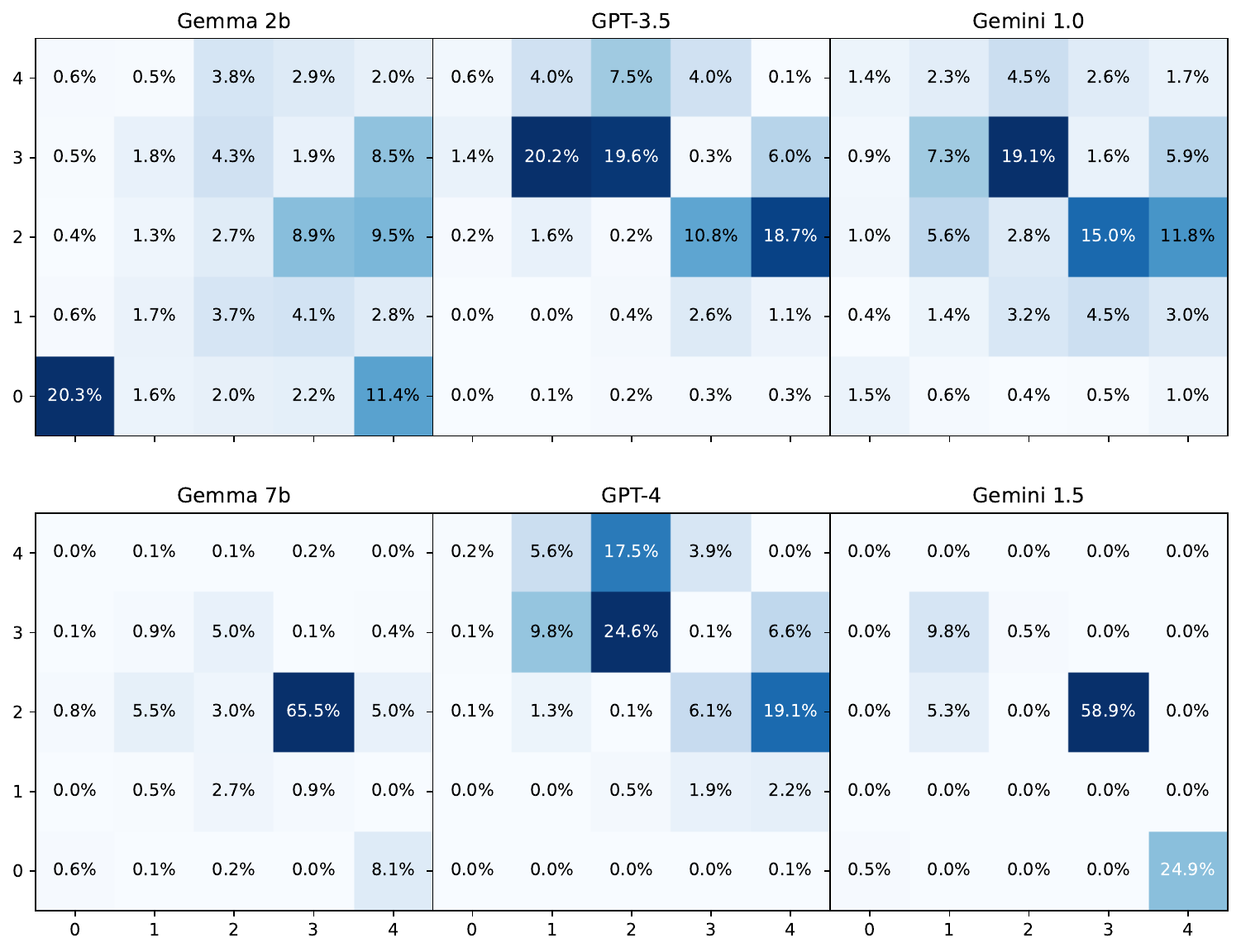}
    \caption{Visualization of the distribution of 1000 random grid locations generated by various LLMs.}
    \label{fig:reward_generation}
\end{figure}

\subsection{Foundation World Models} \label{sec:fwm_experiments}
\paragraph{Experimental setup.}
We examine the ability of LLMs to simulate transition functions, reward functions, and stochastic elements with a set of prompts designed to evaluate these components independently. 
First, we analyze the effect of abstraction by comparing $T$ and $T_\text{minimal}$, which query the model with a detailed and a minimalist description of the transition function, respectively. 
Then, we analyze the effect of adding additional operations by also querying the LLMs for the reward function in both variants, i.e. $T+R$ and $T_\text{minimal}+R$. 
We report the accuracy of the next state predictions, i.e. the model's predictions compared to the true environment outcomes across the entire state-action space. 
We consider a grid world where $n=5$, meaning we predict $|\mathcal{S}|\cdot |\mathcal{A}| = 5^2 \cdot 4 = 100$ transitions for each of the model-prompt combinations.
Additionally, to evaluate the ability of FWMs to facilitate stochastic elements, we challenge each LLM to generate 1000 random grid world (reward) locations.
Additional results on larger grid sizes can be found in Appendix \ref{app:larger_grid} showing that increasing the grid size does not impact the accuracy of the models.

\paragraph{Deterministic results.}
Figure \ref{fig:sim_combined} shows the simulation accuracy and visualizes the corresponding errors across all prompt-model combinations.
The data shows a clear correlation between model capacity and the ability to accurately simulate the environment dynamics. 
There is a noticeable decline in accuracy between detailed $T$ and minimalist prompts $T_\text{minimal}$.
This indicates that inferring transition rules from prior knowledge of grid world dynamics is more challenging than logical reasoning on explicit mathematical constraints.
However, we see that the decline is considerably smaller for the models of greater size. 
Similarly, adding the reward function induces a drop in performance in the medium and small models, but it does not significantly affect the large models.
The visualization of the location of errors in the grid shows that the models particularly struggle with transitions around the edges of the grid.
While the LLMs mostly respect the constraint that actions have no effect when they attempt to move the agent outside the boundaries, they sometimes incorrectly apply this restriction to other actions taken in those locations.
For large models, this happens with $T_\text{minimal}$ but not with $T$ meaning that they can handle all mathematical constraints involved when clearly specified.
In other cases, the models increment or decrement the wrong coordinate.
Additional results that can be found in Appendix \ref{app:larger_grid} show that increasing the grid size does not impact the accuracy of the models.
In summary, these results indicate a positive trend where the capacity of a model significantly influences its ability to accurately simulate environment dynamics and handle more complex prompts.

\paragraph{Stochastic results.} 
Figure \ref{fig:reward_generation} shows the density of the sampled reward locations by the LLMs.
Although no model achieves a distribution that aligns closely with the intended uniform distribution across grid locations, the smaller versions of the same LLMs technology (Gemma 2b, GPT 3.5, and Gemini 1.0) can sample each location at least once. Conversely, Appendix \ref{app:non-uniform} shows that larger models can simulate binary distributions more accurately.
This suggests that smaller models, which are characterized by a larger output variance, can handle larger sample spaces better compared to the larger models, and vice versa.

\subsection{Decision-Making} \label{sec:fa_experiments}
\paragraph{Experimental setup.} To evaluate the integration of FAs and FWMs in the decision-making process of the reinforcement learning framework, we evaluate the two strategies introduced in Section \ref{sec:fa} in both deterministic and stochastic settings. 
For the FWM-based agents, we choose the best-performing FWM from the simulation experiments, i.e. GPT-4 with the $T + R$ prompt.
We pre-train a TRPO agent for the deterministic setting and a RecurrentPPO agent for the stochastic setting, after which their policies are fine-tuned in the true environment.
This allows us to examine the consistency between the simulation and the true environment in both training and testing settings.
In the stochastic setting, the FWM is also asked to generate a random reward location of the grid before the start of each simulated episode.
As a baseline comparison, we also train the agents from scratch in the true environment. 
Both pre-training and true environment interactions are done for 1500 and 1e6 steps for the deterministic and stochastic setting, respectively. 
The agents are evaluated every 125 environment steps with a single episode (with a random reward location in the stochastic setting) across 5 random seeds.
For the FAs, we evaluate all LLMs with the three different prompt approaches as described in \ref{sec:fa}.
Each model-prompt combination is evaluated for 100 independent episodes with $\tau=0$. 

\begin{figure}[t]
    \centering
    \includegraphics[width=\linewidth]{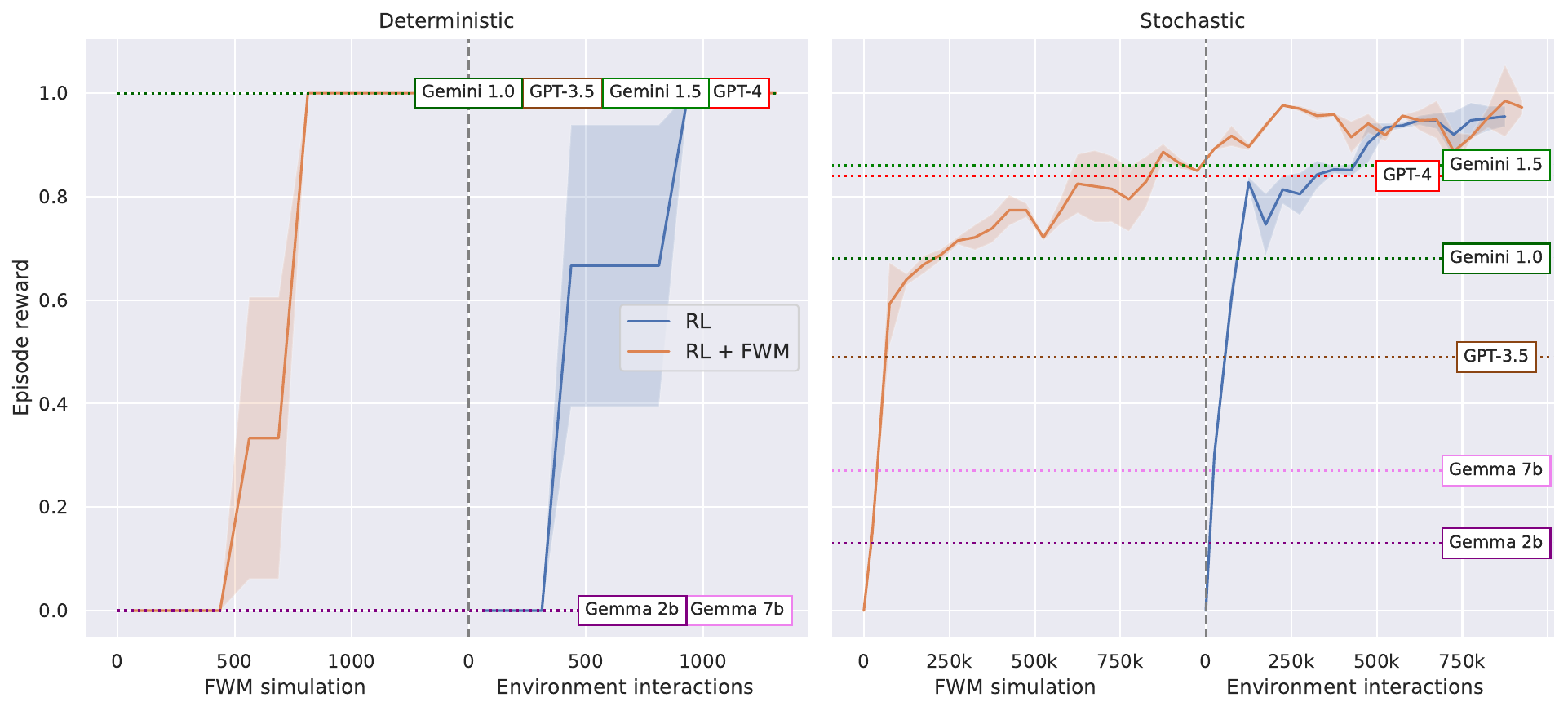}
    \caption{Decision-making results for the deterministic (left) and stochastic (right) setting. The learning curves are visible for the reinforcement learning agents learning from scratch (RL) and the FWM-based agent that first learns through world model interactions (RL + FWM). For each FA the performance of the best prompt approach is reported as a horizontal bar.}
    \label{fig:decision-making-curves}
\end{figure}

\paragraph{Deterministic results.}  
The results for the deterministic setting can be viewed on the left Figure \ref{fig:decision-making-curves} (see Appendix for the numerical results of all FA agents \ref{app:fa_results}). 
We observe that the Gemma models never manage to reach the reward location as they consistently get stuck in the first row or column of the grid. 
However, the GPT- and Gemini-based FA agents excel in this environment regardless of the prompt strategy used. 
Despite having explicit information about the reward location, the reinforcement learning agents need several hundred environment interactions to learn the optimal policy as they are not able to relate the reward location in the observation with the objective of the problem.
In contrast, the reasoning abilities of LLMs allow them to instantly solve the problem. 
This demonstrates that for problems with clear goals and information, an LLM agent can perform extremely well, allowing one to avoid the expensive trial-and-error process inherent in reinforcement learning agents. 
However, the results do show that the TRPO agent is able to learn the optimal policy solely through interacting with the GPT-4 FWM.
When fine-tuning the policy in the true grid world we also observe no further perturbations in the learning curve. 
% Combined with our simulation results from Section \ref{sec:fwm_experiments}, this suggests there is promising potential for foundation models to serve as world models as their capacity increases across future generations.

\begin{figure}[t]
    \centering
    \includegraphics[width=0.95\linewidth]{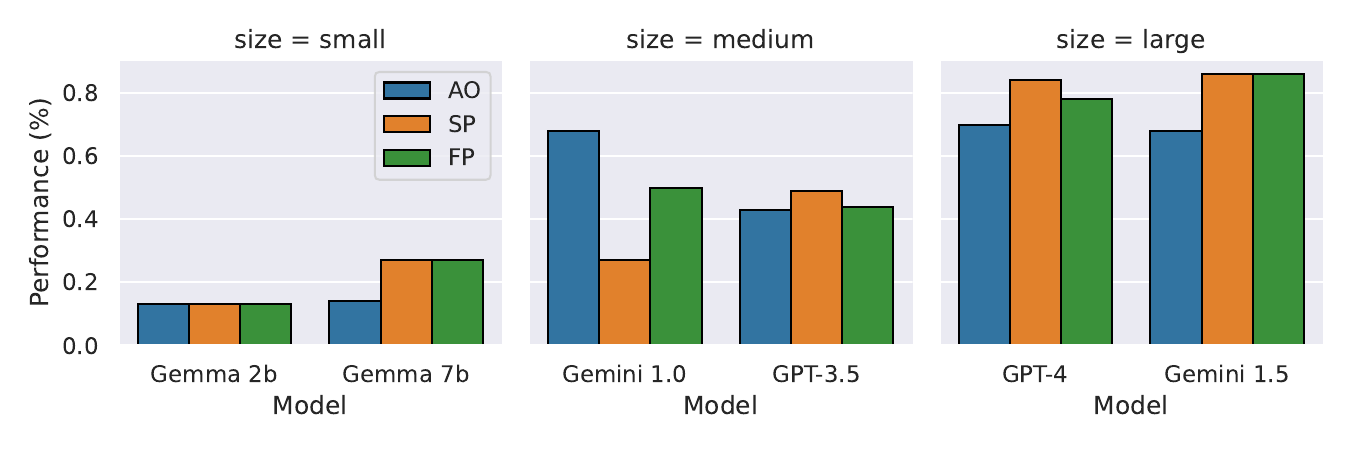}
    \caption{Performances for the FA with prompt strategies Action Only (AO), Simple Plan (SP), and Focused Plan (FP) for the stochastic setting where the reward location is randomized and unknown.}
    \label{fig:fa_stoch_perf}
\end{figure}

\paragraph{Stochastic results.} 
The decision-making performances in the stochastic setting can be found on the right in Figure \ref{fig:decision-making-curves}.
We observe that the Gemma models perform slightly better in this setting, as the randomly generated reward locations may appear in the initial row or column they get fixated on.
We also observe that the FA agents based on larger models struggle significantly more than in the deterministic setting. 
Figure \ref{fig:fa_stoch_perf} shows the individual performances per prompt strategy in the stochastic setting. 
Regardless of prompt strategy or model capacity, none of the agents can consistently search the entire grid within the time limit (see Appendix \ref{app:trajectories} for example trajectories). 
However, we do see significant jumps in performance as the model capacity increases, since GPT-4 and Gemini 1.5 yield policies that search the majority of the grid.
Moreover, in contrast to the smaller models, the larger models seem to benefit significantly from being incentivized to generate plans and utilize their memory through the SP and FP instructions. 
This suggests that the enhanced reasoning abilities resulting from increased model capacity and more sophistication show potential for FAs to dynamically adapt and plan in increasingly complex problems.
Furthermore, although we saw that contemporary FWMs may not be able to facilitate a uniform distribution over possible grid locations, the GPT-4 FWM is still able to provide a sufficient variation of reward locations for the agent to learn a policy that searches the majority of the grid. 
The policy learned by the RecurrentPPO agent smoothly transfers to the true grid environment where it quickly learns a policy that systematically searches the entire grid for the reward, yielding substantially better sample efficiency compared to training from scratch.
In summary, the results show that planning agents are likely to remain preferable in the foreseeable future for tasks that require more sophisticated planning and uncertainty handling, but FWMs can help improve sample efficiency. 

%% file: sections/6_conclusion.tex
\section{Conclusion} \label{sec:conclusion}
In this paper, we made a case in favor of directly integrating foundation models within the reinforcement learning framework.
While previous works focused on using foundation models for simulation and decision-making with intermediate abstractions, we showed that they have promising potential for simulating environment dynamics and low-level action selection.
Our experiments focus on a variety of LLMs and highlight that they can significantly improve sample efficiency in reinforcement learning tasks.

% FWMs simulation result
First, we thoroughly evaluated the ability of the current generation of LLMs to simulate the environment dynamics of a family of grid worlds. 
We found that large models perform exceptionally well at simulating these environments, regardless of the level of detail provided to them about the environment in the instruction prompts. 
Additionally, we explored the performance of these models in simulating probabilistic elements, where we found that smaller models perform better in larger sampling spaces, whereas larger models can more accurately simulate smaller sampling spaces.

% Decision-making
To understand the potential of the integration of foundation models in the decision-making process, we devised two promising strategies: FWM-based agents and foundation agents (FAs). 
The former strategy couples traditional reinforcement learning agents with an LLM-powered world model to pre-train their policies. 
The latter directly utilizes LLMs as low-level action selectors. 
Our experiments showed that in the deterministic setting, FAs excel at navigating and solving grid worlds when the objective is clearly stated in the prompt and may be preferred over traditional reinforcement learning agents.
However, in the stochastic setting, we found that FAs struggled to solve the problem effectively regardless of the model or prompt used, suggesting reinforcement learning agents are still preferable for complex settings.
However, even though we observed LLMs are limited in facilitating such stochastic elements, we found that the FWM-based agents still achieved substantial sample efficiency improvements over training traditional reinforcement learning agents from scratch. 

% Conclusion
Across the experiments conducted in this paper, we found that larger model capacity is associated with better performance.
Considering the rapid advancements and widespread availability of foundation models in various fields the potential for these technologies in reinforcement learning is significant. Future research could explore optimizing FWMs with environment data, employing language model-driven prompt optimization for FWMs, 
leveraging FWMs for planning while interacting with the environment, and combining FAs with FWMs.
Furthermore, the anticipated development of frame prediction foundation models offers an exciting extension to visual environments.

\section{Limitations}
This work serves as a foundational analysis of FMs as world models and agents. Our experiments are limited to text-based grid-worlds, chosen to isolate core reasoning and simulation abilities before moving to visual or embodied domains. We do not compare against learned-from-scratch world models, as our focus is on the zero-shot capabilities of pre-trained FMs. Finally, our FWM-RL method follows a simple pretrain–finetune scheme; more advanced interleaving approaches, such as Dyna-style architectures, remain an important direction for future work.

%% file: sections/8_appendix.tex
\appendix
\section{Simulating Additional Environment Elements} \label{app:more_elements}
In this section, we discuss two further possible environment elements that pose an interesting challenge for FWMs in terms of simulation. We discuss adding a key to the gridworld environment, which requires the model to take a history into account in section \ref{app:history}, and also discuss stochastic elements that follow a non-uniform binary distribution in Section \ref{app:non-uniform}. 

\subsection{History: Gridworld with Key} \label{app:history}
Describing environment dynamics for an FWM is relatively quick and intuitive as natural language is used. For instance, if we were to add a key to the gridworld environment we would need to add two relatively simple conditions to the prompt: the agent picks up a key when at a certain location, and the agent can only access the reward when it possesses the key:
\begin{lstlisting}[breaklines=true, language=JavaScript]
Imagine a gridworld of size <N> by <N> with an observation system of coordinates, where [0,0] is the bottom left corner.

Determine the next observation after taking action <ACTION> from <OBSERVATION> with the following rules:
        - The x and y values are always non-negative and lower than <N-1>.
        - 'right' means increment x
        - 'left' means decrement x 
        - 'up' means increment y
        - 'down' means decrement y.
        - Actions have no effect when they go outside the boundaries.

There is a key located at <KEY_LOCATION>, if this location was previously visited it was acquired.

Determine which reward condition the next observation meets, and respond the corresponding reward and terminal status:
    - The observation is the same as <REWARD_LOCATION> and the key was acquired - 1 True
    - The observation is not the same as <REWARD_LOCATION> or the key was not acquired - 0 and False

Please format your response without any further tokens as:
...
\end{lstlisting}

An interesting thing to note here is that for the FWM to successfully simulate this environment it will need access to the history of previously visited locations. Thus, to facilitate this we could append all previous responses of the LLM within a given episode to the current prompt. However, for environments with extremely long episode durations, this may not be feasible. Although this can be seen as a limitation of FWMs, the processing speed and context length of LLMs are increasing rapidly as of writing, as the Gemini 1.5 model can already process millions of tokens \citep{reid2024gemini}.

\subsection{Non-Uniform Distributions: Gridworld with Sticky Actions}  \label{app:non-uniform}
In this paper, we considered randomized reward locations as the stochastic element of the environment. Another common stochastic element in reinforcement learning environments is sticky actions. Sticky actions repeat the previously taken action with a probability of $\epsilon$. Rather than generating a location across an $n \times n$ grid, simulating the sticky action requires sampling one of two possibilities from a, usually, non-uniform distribution. This incentivizes an interesting experiment: how do LLMs perform when asked to simulate non-uniform distributions? We investigate this with a simple experiment using GPT-3.5 and GPT-4 with the following prompt template:
\begin{lstlisting}[breaklines=true, language=JavaScript]
Respond with 1 with a probability of <P_1> and respond with 0 with a probability of <P_2> with no further tokens.
\end{lstlisting}
We experiment with $p_1 \in [0.6, 0.8, 0.9]$ and $p_2 = 1 - p_1$, and prompt the model 1000 times per combination with $\tau = 1.8$.

\begin{figure}[h]
    \centering
    \includegraphics[width=0.8\linewidth]{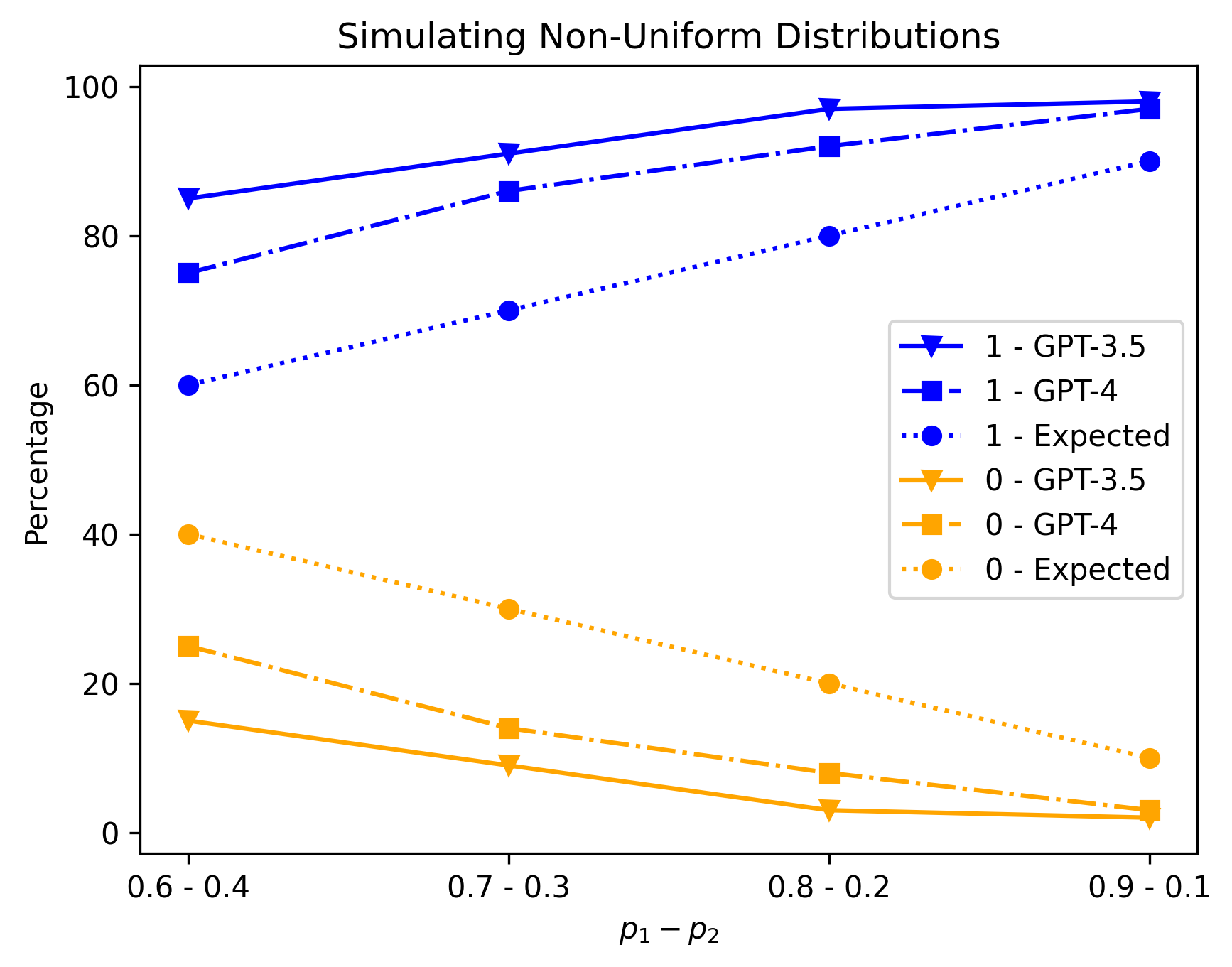}
    \caption{Non-uniform distribution experiment results for GPT-3.5 and GPT-4. The x-axis shows the four probability distributions that were requested. The blue lines represent the frequencies of element '1' belonging to $p_1$, and the orange lines represent the frequencies of element '0' belonging to $p_2$. The triangle markers belong to GPT-3.5, the squares to GPT-4, and the circles to the true expected frequencies.}
    \label{fig:sticky_plot}
\end{figure}

As can be seen in Table \ref{tab:sticky_experiments} and Figure \ref{fig:sticky_plot}, it is clear that the LLM tends to overestimate the likelihood of choosing the higher probability $p_1$. However, despite not exactly matching the requested distributions, the resulting distributions yielded by the LLMs do follow a progressively increasing discrepancy trend similar to the expected true distributions (dotted lines in the figure). This indicates that the LLMs are able to adjust the likelihood of sampling each element such that the discrepancy increases accordingly. Furthermore, GPT-4 yields discrepancies that match the true distributions significantly more than GPT-3.5 does. This suggests that larger models are increasingly better at simulating stochastic elements with small sample spaces. 

\begin{table}[h]
    \caption{Non-uniform distribution experiment numerical results for GPT-3.5 and GPT-4 computed from 1000 samples.}
    \centering
        \begin{tabular}{c|cc|cc}
        \toprule
        $p_1 - p_2$ & 1 - GPT3.5 & 0 - GPT3.5 & 1 - GPT4 & 0 - GPT4  \\
        \midrule
        $0.6 - 0.4$ &  85\% & 15\% &  75\% & 25\% \\
        $0.7 - 0.3$ &  91\% & 9\% &  86\% & 14\%\\
        $0.8 - 0.2$ &  97\% & 3\% & 92 \% & 8\% \\
        $0.9 - 0.1$ &  98\% & 2\% & 97\% & 3\% \\
        \bottomrule
        \end{tabular}
    \label{tab:sticky_experiments}
\end{table}

\clearpage

\section{Implementation}
\subsection{Reinforcement Learning Agents} \label{app:hyperparams}
As reinforcement learning agents the Trust Region Policy Optimization (TRPO) \citep{schulman2015trust} and Recurrent Proximal Policy Optimization (RecurrentPPO) \citep{schulman2017proximal} were used. We used the default implementations provided by Stable Baslines 3 \citep{stable-baselines3}. We used the default hyperparameters, except for the update frequency that was changed to 125, and the learning rate for RecurrentPPO was decreased to 3e-5. 

\subsection{Foundation World Models} \label{app:sim_prompts}
In order to construct a connection between a reinforcement learning agent and a FWM, we ask the LLM to format its response as follows:
\begin{lstlisting}[breaklines=true, language=JavaScript]
Please format your response without any further tokens as:
[x, y, <n>-1, <n>-1], reward, terminal
\end{lstlisting}
or in the stochastic setting:
\begin{lstlisting}[breaklines=true, language=JavaScript]
Please format your response without any further tokens as:
[x, y], reward, terminal
\end{lstlisting}
The responded string can conveniently be converted to a tuple that is typically returned by reinforcement learning environments. In the case of reward generation with $\tau = 1.8$, the LLM could respond with nonsensical content occasionally, for which we were required to employ regex expressions to extract the generated reward location. As the transition functions of the environments are deterministic, we also used caching to reduce the number of required prompts for the FWM. 

The full prompt templates used for $T$, $T+R$, $T_\text{minimal}$, and $T_\text{minimal} + R$ can be found in Listing \ref{p}, Listing \ref{pr}, Listing \ref{pmin}, and Listing \ref{pminr}, respectively.

\begin{lstlisting}[breaklines=true, language=JavaScript, caption=Prompt template for $T$., label={p}]
Imagine a gridworld of size <n> by <n> with an observation system of coordinates, where [0,0] is the bottom left corner. 

Determine the next observation after taking action <ACTION> from <OBSERVATION> with the following rules:
        - The x and y values are always non-negative and lower than <n-1>.
        - 'right' means increment x
        - 'left' means decrement x 
        - 'up' means increment y
        - 'down' means decrement y.
        - Actions have no effect when they go outside the boundaries.

Please format your response without any further tokens as:
...
\end{lstlisting}

\begin{lstlisting}[breaklines=true, language=JavaScript, caption=Prompt template for $T + R$., label={pr}]
Imagine a gridworld of size <n> by <n> with an observation system of coordinates, where [0,0] is the bottom left corner. 

Determine the next observation after taking action <ACTION> from <OBSERVATION> with the following rules:
        - The x and y values are always non-negative and lower than <n-1>.
        - 'right' means increment x
        - 'left' means decrement x 
        - 'up' means increment y
        - 'down' means decrement y.
        - Actions have no effect when they go outside the boundaries.

Determine which reward condition the next observation meets, and respond the corresponding reward and terminal status:
    - The observation is the same as <REWARD LOCATION> - 1 and True
    - The observation is not the same as <REWARD LOCATION> - 0 and False

Please format your response without any further tokens as:
...
\end{lstlisting}

\begin{lstlisting}[breaklines=true, language=JavaScript, caption=Prompt template for $T_\text{minimal} + R$., label={pmin}]
Imagine a gridworld of size <n> by <n> with an observation system of coordinates. 

Determine the next observation after taking action <ACTION> from <OBSERVATION> with the following rules:
    - The location [0,0] is the bottom-left corner, and [<n-1>,<n-1>] it the top-right corner.

Please format your response without any further tokens as:
...
\end{lstlisting}

\begin{lstlisting}[breaklines=true, language=JavaScript, caption=Prompt template for $T_\text{minimal} + R$., label={pminr}]
Imagine a gridworld of size <n> by <n> with an observation system of coordinates. 

Determine the next observation after taking action <ACTION> from <OBSERVATION> with the following rules:
    - The location [0,0] is the bottom-left corner, and [<n-1>,<n-1>] it the top-right corner.

Determine which reward condition the next observation meets, and respond the corresponding reward and terminal status:
    - The observation is the same as <REWARD LOCATION> - 1 and True
    - The observation is not the same as <REWARD LOCATION> - 0 and False

Please format your response without any further tokens as:
...
\end{lstlisting}

\subsection{Foundation Agents} \label{app:prompt_templates_llms}
The prompt templates used for the deterministic and stochastic LLM agents are almost identical. While the deterministic agent receives the exact position where the reward is located, the stochastic agent is only told the following: \textit{Your goal is to reach the reward located at a random coordinate as quickly as possible.} See Listing \ref{listing_llm_agent_prompt} for the full prompt used by the deterministic LLM agent. The agent's memory is filled as it interacts with the game. Whenever an action is executed, we add the following line to the memory: \textit{Executed <ACTION> at <LOCATION> resulting in <NEW LOCATION> and no reward.} Additionally, as seen in Listing \ref{listing_simple_plan}, whenever an agent chooses an action, it outputs a plan representing its thoughts. We add each plan to the memory as well.

\begin{lstlisting}[breaklines=true, language=JavaScript, caption=The prompt template for the deterministic LLM agent., label={listing_llm_agent_prompt}]
**Context**
You are an agent in a <n>x<n> grid.
The bottom left corner is at <BOTTOM LEFT>, top left at [0, n-1], top right at [<n-1, n-1>], and bottom right at [<n-1>, 0].
The x-axis increases as you move rightward, and the y axis increases as you move upwards. 
Your goal is to reach the reward located at a coordinate [<n-1>,<n-1>] (the top-right corner).

**Memory**
[...]

**Observation**
Your current location is <OBSERVATION>

**Available Actions**
up
right
down
left

**Task**
Choose an action from the given list of actions. Output your response using the following JSOn format and do not use markdown.
<OUTPUT FORMAT>
\end{lstlisting}

The three prompting approaches described in section \ref{sec:fa_experiments} are implemented using different JSON output formats. Note that since the action-only agent outputs no plan, its memory will only contain the results of the executed actions. Note that in the stochastic setting, we do not mention that the reward is located in the top-right corner.

\begin{lstlisting}[breaklines=true, language=JavaScript, caption=The output format used by the \textbf{Action Only} agent.]
{
    "action": "<The action you want to take>"
}
\end{lstlisting}

\begin{lstlisting}[breaklines=true, language=JavaScript, caption=The output format used by the \textbf{Simple Plan} agent., label={listing_simple_plan}]
{
    "plan": "<Think about what you want to do next to fulfill your goal.>",
    "action": "<The action you want to take>"
}
\end{lstlisting}

\begin{lstlisting}[breaklines=true, language=JavaScript, caption=The output format used by the \textbf{Focused Plan} agent.]
{
    "plan": "<Use your memory to determine which position you want to go next.>",
    "analysis": "<Using your plan analyze which action is best to efficiently reach the position.>",
    "action": "<The action you want to take>"
}
\end{lstlisting}

\newpage \newpage

\section{Additional Simulation Results}  \label{app:all_generation_results}
\subsection{Numerical Results}
Table \ref{tab:accuracies} shows the numerical values for the simulation of the transition function experiments performed in Section \ref{sec:fwm_experiments}. 

\begin{table}[h] 
    \centering
    \footnotesize
    \caption{Simulation accuracy results for all prompt-model combinations. Prompt $T$ queries the transition function with detailed descriptions, $T + R$ queries both the transition and reward function with detailed descriptions, and prompt $T_\text{minimal}$ and $T_\text{minimal} + R$ the former prompts with minimal transition function descriptions.}
    \begin{tabular}{c|cccccc}
    \toprule
    \textbf{Prompt} & \textbf{Gemma 2b} & \textbf{Gemma 7b} & \textbf{Gemini 1.0} & \textbf{GPT-3.5} & \textbf{GPT-4} & \textbf{Gemini 1.5}\\
    \midrule
    $T$ & 13\% & 40\% & 84\% & 98\% & 100\% & 100\% \\  
    $T + R$ & 12\% & 32\% & 82\% & 94\% & 100\% & 100\% \\
    $T_\text{minimal}$ & 22\% & 28\% & 58\% & 89\% & 96\% & 97\% \\
    $T_\text{minimal} + R$ & 4\% & 31\% & 75\% & 88\% & 96\% & 96\% \\
    \bottomrule
    \end{tabular}
    \vskip -0.1in
    \label{tab:accuracies}
\end{table}

\subsection{Larger Grids} \label{app:larger_grid}
\begin{figure}[h]
    \centering
    \includegraphics[width=\linewidth]{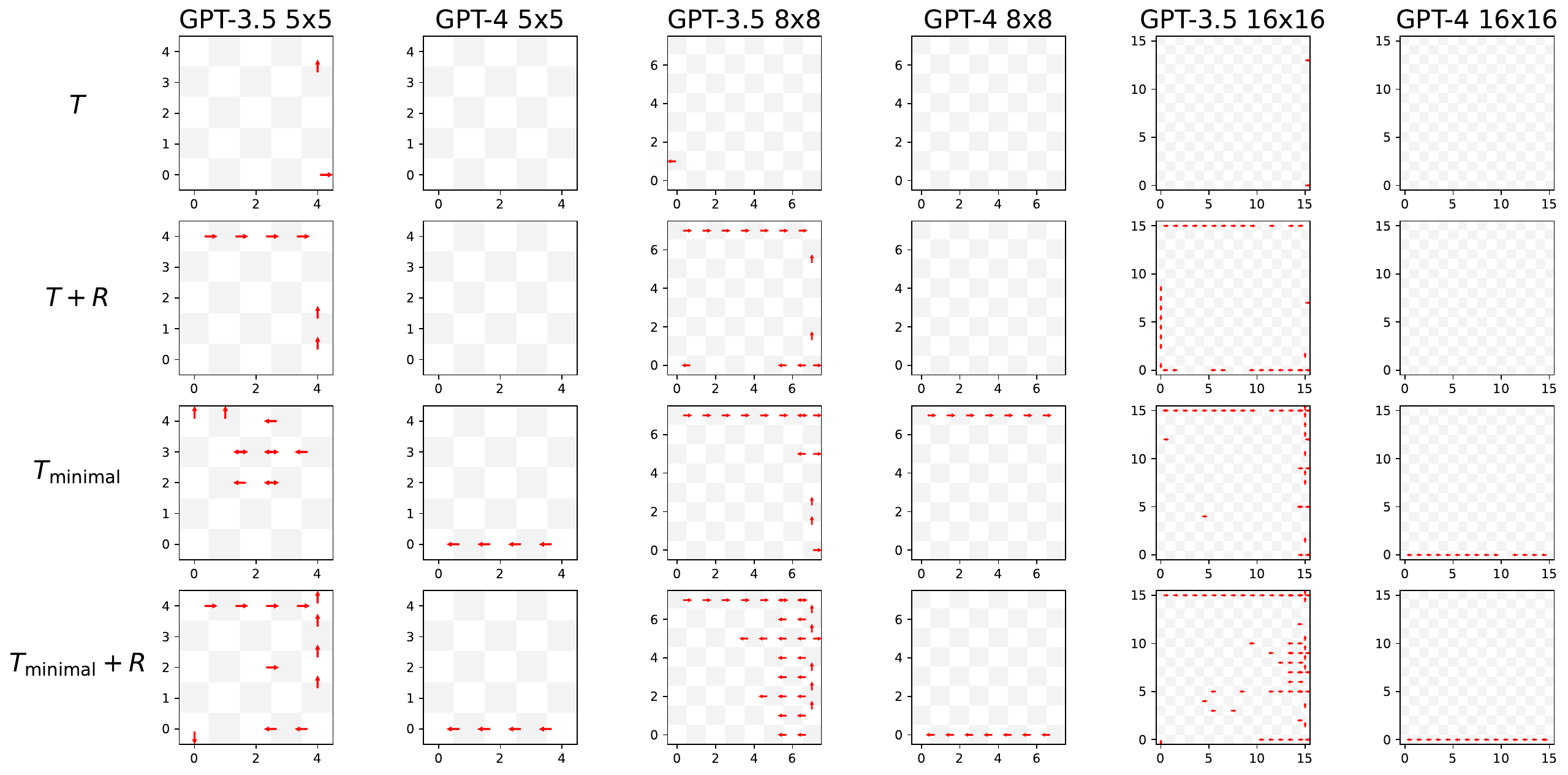}
    \caption{Visualization of the errors for GPT-3.5 and GPT-4 for grid sizes of 8x8 and 16x16.}
    \label{fig:simulation_errors_8x8_16x16}
\end{figure}

Table \ref{tab:accuracies_larger_grids} shows the simulation accuracy for the GPT-3.5 and GPT-4 models with grid sizes $n \in [5, 8, 16]$. Figure \ref{fig:simulation_errors_8x8_16x16} shows the errors made by the models. As can be observed, the models remain consistent in terms of accuracy despite the grid size increases. 

\begin{table}[h] 
    \centering
    \footnotesize
    \caption{Simulation accuracy results for the GPT models with varying grid sizes. Prompt $T$ queries the transition function with detailed descriptions, $T + R$ queries both the transition and reward function with detailed descriptions, and prompt $T_\text{minimal}$ and $T_\text{minimal} + R$ the former prompts with minimal transition function descriptions.}
    \begin{tabular}{c|cccccc}
    \toprule
    \textbf{Prompt} & \textbf{GPT-3.5 5x5} & \textbf{GPT-4 5x5} & \textbf{GPT-3.5 8x8} & \textbf{GPT-4 8x8} & \textbf{GPT-3.5 16x16} & \textbf{GPT-4 16x16}\\
    \midrule
    $T$ & 98\% & 100\% & 99\% & 100\% & 99\% & 100\% \\  
    $T + R$ & 94\% & 100\% & 95\% & 100\% & 97\% & 100\% \\  
    $T_\text{minimal} $ & 89\% & 96\% & 95\% & 97\% & 96\% & 99\% \\  
    $T_\text{minimal} + R$ & 88\% & 96\% & 88\% & 97\% & 94\% & 98\% \\  
    \bottomrule
    \end{tabular}
    \vskip -0.1in
    \label{tab:accuracies_larger_grids}
\end{table}

\clearpage

\section{Additional Foundation Agent Results} \label{app:fa_results}
\begin{figure}[h]
    \centering
    \includegraphics[width=\linewidth]{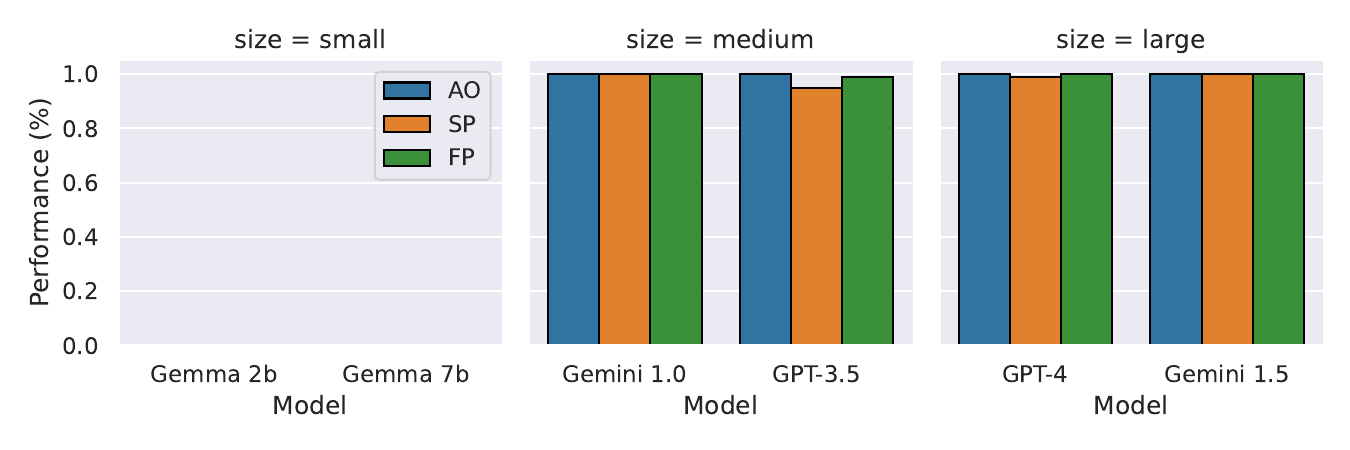}
    \caption{Performances for the FA with prompt strategies Action Only (AO), Simple Plan (SP), and Focused Plan (FP) for the deterministic setting with a fixed and known reward location.}
    \label{fig:fa_det_perf}
\end{figure}

The full list of numerical results for the FA performances from the experiments in Section \ref{sec:fa_experiments} can be found in Table \ref{tab:agent_performances}. Further, a visualization of the deterministic gridworld performances can be found in Figure \ref{fig:fa_det_perf}. 

\begin{table}[h]
    \caption{LLM Agent performances for an empty $5\times5$ grid with a fixed reward location and random reward location, averaged over 100 episodes.}
    \centering
        \begin{tabular}{lcc}
        \toprule
        \textbf{Model} & \textbf{Fixed Reward} & \textbf{Random Reward} \\
        \midrule
        Gemma 2b (AO) & 0\% &  13\%\\
        Gemma 2b (SP) & 0\% &  13\%\\
        Gemma 2b (FP) & 0\% &  13\% \\
        \hline
        Gemma 7b (AO) & 0\% &  14\% \\
        Gemma 7b (SP) & 0\% &  27\% \\
        Gemma 7b (FP) & 0\% &  27\%\\
                \hline

        Gemini 1.0 (AO) & 100\%  & 68\% \\
        Gemini 1.0 (SP) & 100\% & 27\%\\
        Gemini 1.0 (FP) & 100\% & 50\% \\
                \hline
                
        GPT-3.5 (AO) & 100\% & 43\% \\
        GPT-3.5 (SP) & 95\% & 49\%  \\
        GPT-3.5 (FP) & 99\% &  44\%\\
                \hline

        GPT-4 (AO) & 100\% & 70\% \\
        GPT-4 (SP) & 99\% & 84\% \\
        GPT-4 (FP) & 100\% & 78\% \\
                \hline

        Gemini 1.5 (AO) & 100\% & 68\% \\
        Gemini 1.5 (SP) & 100\% & 86\% \\
        Gemini 1.5 (FP) & 100\% & 86\% \\
        \bottomrule
        \end{tabular}
    \label{tab:agent_performances}
\end{table}

% \begin{figure}[h]
%     \centering
%     \includegraphics[width=\linewidth]{images/fa_performances_groups_stoch.png}
%     \caption{Performances for the FA with prompt strategies Action Only (AO), Simple Plan (SP), and Focused Plan (FP) for the stochastic setting with a randomized and unknown reward location.}
%     \label{fig:fa_stoch_perf}
% \end{figure}
\clearpage

\section{Decision-Making Trajectories} \label{app:trajectories}
\begin{figure}[h]
    \centering
    \includegraphics[width=0.85\linewidth]{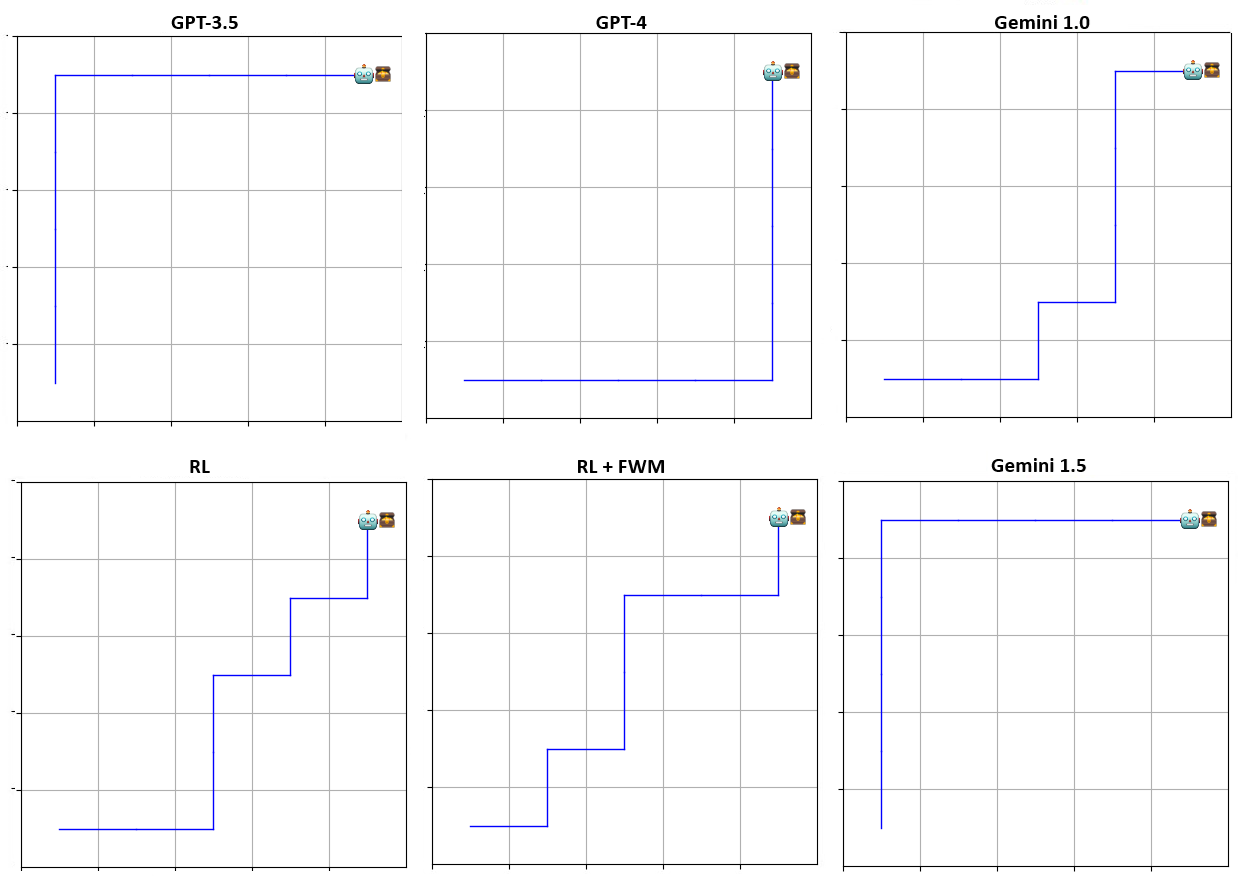}
    \caption{Example trajectories of various decision-making agents in the deterministic setting.}
    \label{fig:trajectories_det}
\end{figure}

\begin{figure}[h]
    \centering
    \includegraphics[width=0.85\linewidth]{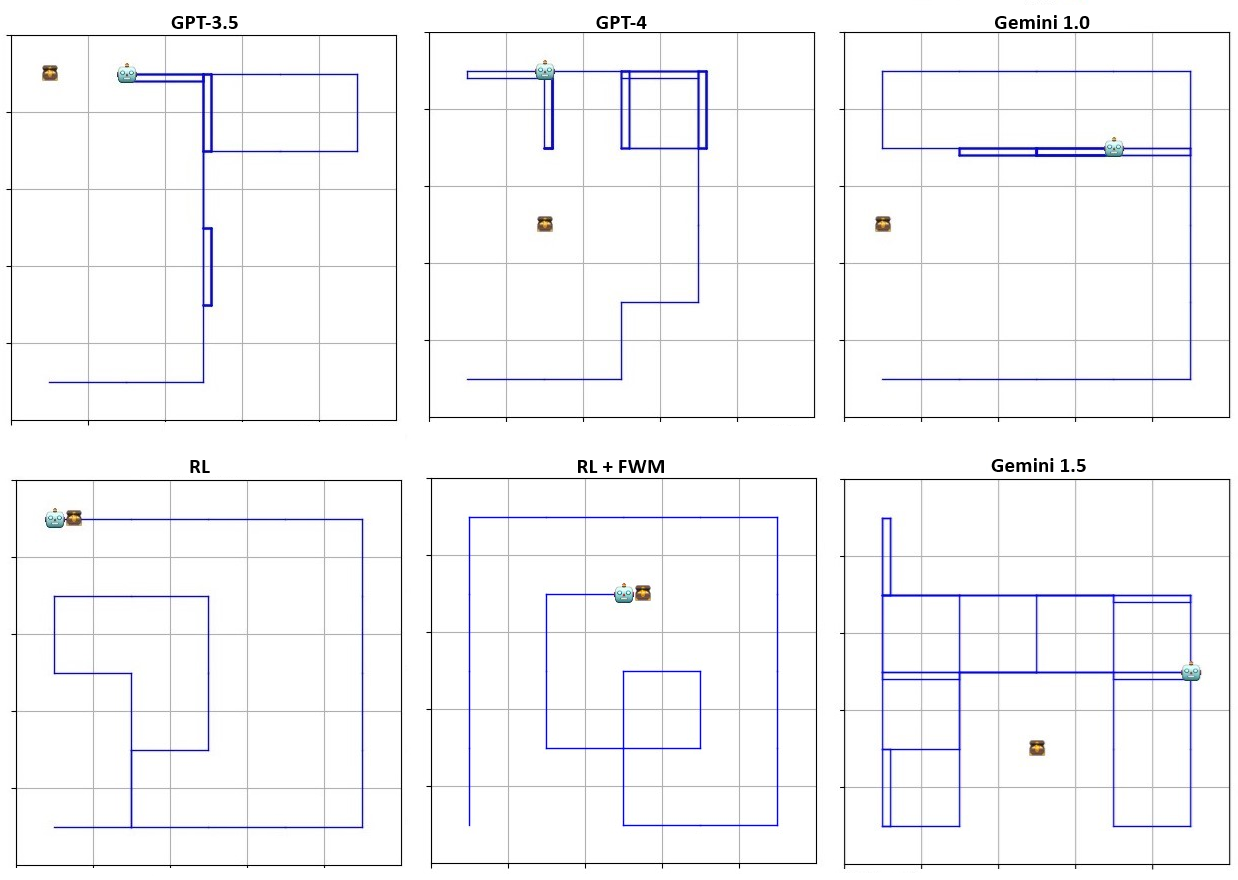}
    \caption{Example trajectories of various decision-making agents in the stochastic setting.}
    \label{fig:trajectories}
\end{figure}

In Figure \ref{fig:trajectories_det} and Figure \ref{fig:trajectories} example trajectories of the various decision-making agents in the deterministic and stochastic setting can be found, respectively. Although all agents successfully learn to find the rewards in various ways in the deterministic setting, in the stochastic setting the FAs tend to get stuck in certain regions of the grid. The reinforcement learning agents learn to successfully search the grid systematically in the stochastic setting. Note that there are no visualizations of the Gemma models as they simply got stuck in the first row or column.

\clearpage

\section{Compute} \label{app:compute}
We use OpenAI’s APIs for GPT-3.5 and GPT-4 and Google Studio's APIs for Gemini 1.0 and Gemini 1.5. For Gemma 2b and Gemma 7b, we made use of the Pytorch implementation provided by Google AI for Developers. We ran these Gemma models on a single NVIDIA A100 GPU, as 8GB+ RAM on GPU for the 2b and 24GB+ RAM on GPU for 7b are recommended. For training the PPO and TRPO agents we also used a single NVIDIA A100 GPU in all environments, but when using the API-based models this could also be run on a CPU.

For the GPT models, we used 'gpt-4-0613' and 'gpt-3.5-turbo-0613' cutoffs, which cost US\$ 0,50 / 1M tokens US\$ 1,50 / 1M tokens and US\$ 30,00 / 1M tokens US\$ 60,00 / 1M tokens, respectively, as of writing. The Gemini and Gemma models can be used freely as of writing, although the Gemini models have a relatively low limit for queries per minute. For the Gemini models we used 'gemini-1.0-pro-001' and 'gemini-1.5-pro'. For the Gemma models, we used 'Gemma-2b-it' and 'Gemma-7b-it' without quantization. 

The runtime for a RecurrentPPO agent in the stochastic setting using a GPT-4 FWM with caching and an NVIDIA A100 GPU for 1e6 environment steps, followed by 1e6 fine-tuning steps in the true gridworld, takes under 2 hours.

\section{Societal Impact} \label{app:societal_impact}
The methodologies proposed in this paper are primarily benign in isolation but, crucially, when future LLMs and other foundation models are leveraged as simulators or action selectors in sequential decision-making problems for real-world applications, specific cautionary measures are necessary. For instance, the use of these models in dynamic, real-time decision-making environments such as autonomous driving would introduce significant ethical and societal challenges. To address these risks rigorous validation processes to ensure that models behave as intended in varied and unforeseen circumstances should be used, similar to the extensive experimentation performed in this paper where we pinpoint and investigate the errors made by these models. Conversely, the ability of foundation models to simulate environments has major positive potential as this would enable accessible simulation of real-world scenarios in which decision-making agents can extensively be optimized and evaluated.

%% file: neurips_2024.bbl
\begin{thebibliography}{42}
\providecommand{\natexlab}[1]{#1}
\providecommand{\url}[1]{\texttt{#1}}
\expandafter\ifx\csname urlstyle\endcsname\relax
  \providecommand{\doi}[1]{doi: #1}\else
  \providecommand{\doi}{doi: \begingroup \urlstyle{rm}\Url}\fi

\bibitem[Achiam et~al.(2023)Achiam, Adler, Agarwal, Ahmad, Akkaya, Aleman, Almeida, Altenschmidt, Altman, Anadkat, et~al.]{achiam2023gpt}
Achiam, J., Adler, S., Agarwal, S., Ahmad, L., Akkaya, I., Aleman, F.~L., Almeida, D., Altenschmidt, J., Altman, S., Anadkat, S., et~al.
\newblock Gpt-4 technical report.
\newblock \emph{arXiv preprint arXiv:2303.08774}, 2023.

\bibitem[Berner et~al.(2019)Berner, Brockman, Chan, Cheung, Debiak, Dennison, Farhi, Fischer, Hashme, Hesse, et~al.]{berner2019dota}
Berner, C., Brockman, G., Chan, B., Cheung, V., Debiak, P., Dennison, C., Farhi, D., Fischer, Q., Hashme, S., Hesse, C., et~al.
\newblock Dota 2 with large scale deep reinforcement learning.
\newblock \emph{arXiv preprint}, 2019.

\bibitem[Besta et~al.(2024)]{besta2024graph}
Besta, M. et~al.
\newblock Graph of thoughts: Solving elaborate problems with large language models.
\newblock \emph{arXiv preprint arXiv:2402.15732}, 2024.

\bibitem[Brown et~al.(2020)Brown, Mann, Ryder, Subbiah, Kaplan, Dhariwal, Neelakantan, Shyam, Sastry, Askell, Agarwal, Herbert-Voss, Krueger, Henighan, Child, Ramesh, Ziegler, Wu, Winter, Hesse, Chen, Sigler, Litwin, Gray, Chess, Clark, Berner, McCandlish, Radford, Sutskever, and Amodei]{brown2020language}
Brown, T., Mann, B., Ryder, N., Subbiah, M., Kaplan, J.~D., Dhariwal, P., Neelakantan, A., Shyam, P., Sastry, G., Askell, A., Agarwal, S., Herbert-Voss, A., Krueger, G., Henighan, T., Child, R., Ramesh, A., Ziegler, D., Wu, J., Winter, C., Hesse, C., Chen, M., Sigler, E., Litwin, M., Gray, S., Chess, B., Clark, J., Berner, C., McCandlish, S., Radford, A., Sutskever, I., and Amodei, D.
\newblock Language models are few-shot learners.
\newblock In Larochelle, H., Ranzato, M., Hadsell, R., Balcan, M., and Lin, H. (eds.), \emph{Advances in Neural Information Processing Systems}, volume~33, pp.\  1877--1901. Curran Associates, Inc., 2020.
\newblock URL \url{https://proceedings.neurips.cc/paper_files/paper/2020/file/1457c0d6bfcb4967418bfb8ac142f64a-Paper.pdf}.

\bibitem[Chevalier-Boisvert et~al.(2023)Chevalier-Boisvert, Dai, Towers, de~Lazcano, Willems, Lahlou, Pal, Castro, and Terry]{MinigridMiniworld23}
Chevalier-Boisvert, M., Dai, B., Towers, M., de~Lazcano, R., Willems, L., Lahlou, S., Pal, S., Castro, P.~S., and Terry, J.
\newblock Minigrid \& miniworld: Modular \& customizable reinforcement learning environments for goal-oriented tasks.
\newblock \emph{CoRR}, abs/2306.13831, 2023.
\newblock URL \url{https://github.com/Farama-Foundation/Minigrid}.

\bibitem[Chiang et~al.(2024)Chiang, Zheng, Sheng, Angelopoulos, Li, Li, Zhang, Zhu, Jordan, Gonzalez, et~al.]{chiang2024chatbot}
Chiang, W.-L., Zheng, L., Sheng, Y., Angelopoulos, A.~N., Li, T., Li, D., Zhang, H., Zhu, B., Jordan, M., Gonzalez, J.~E., et~al.
\newblock Chatbot arena: An open platform for evaluating llms by human preference.
\newblock \emph{arXiv preprint arXiv:2403.04132}, 2024.

\bibitem[Dainese et~al.(2024)]{dainese2024gifmcts}
Dainese, M. et~al.
\newblock Code world models with gif-mcts.
\newblock In \emph{Advances in Neural Information Processing Systems (NeurIPS)}, 2024.

\bibitem[DeepMind(2025)]{deepmind2025genie3}
DeepMind.
\newblock Genie 3: A new frontier for world models.
\newblock \url{https://deepmind.google/discover/blog/genie-3-a-new-frontier-for-world-models/}, 2025.
\newblock Accessed: 2025-09-02.

\bibitem[Driess et~al.(2023)Driess, Xia, Sajjadi, Lynch, Chowdhery, Ichter, Wahid, Tompson, Vuong, Yu, et~al.]{driess2023palm}
Driess, D., Xia, F., Sajjadi, M.~S., Lynch, C., Chowdhery, A., Ichter, B., Wahid, A., Tompson, J., Vuong, Q., Yu, T., et~al.
\newblock Palm-e: An embodied multimodal language model.
\newblock \emph{arXiv preprint arXiv:2303.03378}, 2023.

\bibitem[Du et~al.(2023)Du, Watkins, Wang, Colas, Darrell, Abbeel, Gupta, and Andreas]{yuqing2023guidingpretraining}
Du, Y., Watkins, O., Wang, Z., Colas, C., Darrell, T., Abbeel, P., Gupta, A., and Andreas, J.
\newblock Guiding pretraining in reinforcement learning with large language models.
\newblock In Krause, A., Brunskill, E., Cho, K., Engelhardt, B., Sabato, S., and Scarlett, J. (eds.), \emph{Proceedings of the 40th International Conference on Machine Learning}, volume 202 of \emph{Proceedings of Machine Learning Research}, pp.\  8657--8677. PMLR, 23--29 Jul 2023.
\newblock URL \url{https://proceedings.mlr.press/v202/du23f.html}.

\bibitem[Guan et~al.(2023)Guan, Valmeekam, Sreedharan, and Kambhampati]{guan2023leveraging}
Guan, L., Valmeekam, K., Sreedharan, S., and Kambhampati, S.
\newblock Leveraging pre-trained large language models to construct and utilize world models for model-based task planning.
\newblock In \emph{Thirty-seventh Conference on Neural Information Processing Systems}, 2023.
\newblock URL \url{https://openreview.net/forum?id=zDbsSscmuj}.

\bibitem[Ha \& Schmidhuber(2018)Ha and Schmidhuber]{Ha2018-jd}
Ha, D. and Schmidhuber, J.
\newblock Recurrent world models facilitate policy evolution.
\newblock In Bengio, S., Wallach, H., Larochelle, H., Grauman, K., Cesa-Bianchi, N., and Garnett, R. (eds.), \emph{Advances in Neural Information Processing Systems}, volume~31. Curran Associates, Inc., 2018.
\newblock URL \url{https://proceedings.neurips.cc/paper_files/paper/2018/file/2de5d16682c3c35007e4e92982f1a2ba-Paper.pdf}.

\bibitem[Hafner et~al.(2023)Hafner, Pasukonis, Ba, and Lillicrap]{hafner2023mastering}
Hafner, D., Pasukonis, J., Ba, J., and Lillicrap, T.
\newblock Mastering diverse domains through world models.
\newblock \emph{arXiv preprint arXiv:2301.04104}, 2023.

\bibitem[Hao et~al.(2023)Hao, Gu, Ma, Hong, Wang, Wang, and Hu]{hao2023reasoning}
Hao, S., Gu, Y., Ma, H., Hong, J.~J., Wang, Z., Wang, D.~Z., and Hu, Z.
\newblock Reasoning with language model is planning with world model.
\newblock In \emph{The 2023 Conference on Empirical Methods in Natural Language Processing}, 2023.
\newblock URL \url{https://openreview.net/forum?id=VTWWvYtF1R}.

\bibitem[Hao* et~al.(2024)Hao*, Gu*, Luo*, Liu, Shao, Wang, Xie, Ma, Samavedhi, Gao, et~al.]{hao2024llm}
Hao*, S., Gu*, Y., Luo*, H., Liu, T., Shao, X., Wang, X., Xie, S., Ma, H., Samavedhi, A., Gao, Q., et~al.
\newblock Llm reasoners: New evaluation, library, and analysis of step-by-step reasoning with large language models.
\newblock \emph{arXiv preprint arXiv:2404.05221}, 2024.

\bibitem[Janner et~al.(2019)Janner, Fu, Zhang, and Levine]{janner2019trust}
Janner, M., Fu, J., Zhang, M., and Levine, S.
\newblock When to trust your model: Model-based policy optimization.
\newblock In Wallach, H., Larochelle, H., Beygelzimer, A., d\textquotesingle Alch\'{e}-Buc, F., Fox, E., and Garnett, R. (eds.), \emph{Advances in Neural Information Processing Systems}, volume~32. Curran Associates, Inc., 2019.
\newblock URL \url{https://proceedings.neurips.cc/paper_files/paper/2019/file/5faf461eff3099671ad63c6f3f094f7f-Paper.pdf}.

\bibitem[Kaiser et~al.(2020)Kaiser, Babaeizadeh, Milos, Osinski, Campbell, Czechowski, Erhan, Finn, Kozakowski, Levine, et~al.]{Kaiser2019-ts}
Kaiser, L., Babaeizadeh, M., Milos, P., Osinski, B., Campbell, R.~H., Czechowski, K., Erhan, D., Finn, C., Kozakowski, P., Levine, S., et~al.
\newblock Model based reinforcement learning for atari.
\newblock In \emph{International Conference on Learning Representations}, 2020.
\newblock URL \url{https://openreview.net/forum?id=S1xCPJHtDB}.

\bibitem[Klissarov et~al.(2024)Klissarov, D'Oro, Sodhani, Raileanu, Bacon, Vincent, Zhang, and Henaff]{klissarov2024motif}
Klissarov, M., D'Oro, P., Sodhani, S., Raileanu, R., Bacon, P.-L., Vincent, P., Zhang, A., and Henaff, M.
\newblock Motif: Intrinsic motivation from artificial intelligence feedback.
\newblock In \emph{The Twelfth International Conference on Learning Representations}, 2024.
\newblock URL \url{https://openreview.net/forum?id=tmBKIecDE9}.

\bibitem[Lin et~al.(2023)Lin, Agia, Migimatsu, Pavone, and Bohg]{lin2023text2motion}
Lin, K., Agia, C., Migimatsu, T., Pavone, M., and Bohg, J.
\newblock Text2motion: from natural language instructions to feasible plans.
\newblock \emph{Autonomous Robots}, Nov 2023.
\newblock ISSN 1573-7527.
\newblock \doi{10.1007/s10514-023-10131-7}.
\newblock URL \url{https://doi.org/10.1007/s10514-023-10131-7}.

\bibitem[Liu et~al.(2023)]{liu2023rafa}
Liu, X. et~al.
\newblock Reason for future, act for now: A simple approach for llm planning.
\newblock In \emph{Proceedings of the 37th AAAI Conference on Artificial Intelligence}, 2023.

\bibitem[Ma et~al.(2024{\natexlab{a}})Ma, Liang, Wang, Huang, Bastani, Jayaraman, Zhu, Fan, and Anandkumar]{eureka}
Ma, Y.~J., Liang, W., Wang, G., Huang, D.-A., Bastani, O., Jayaraman, D., Zhu, Y., Fan, L., and Anandkumar, A.
\newblock Eureka: Human-level reward design via coding large language models.
\newblock In \emph{The Twelfth International Conference on Learning Representations}, 2024{\natexlab{a}}.
\newblock URL \url{https://openreview.net/forum?id=IEduRUO55F}.

\bibitem[Ma et~al.(2024{\natexlab{b}})]{ma2024fme}
Ma, Z. et~al.
\newblock Foundation model exploration: Improving reinforcement learning with vlm-guided exploration.
\newblock In \emph{Proceedings of the International Conference on Learning Representations (ICLR)}, 2024{\natexlab{b}}.

\bibitem[Papamarkou et~al.(2024)Papamarkou, Skoularidou, Palla, Aitchison, Arbel, Dunson, Filippone, Fortuin, Hennig, Hubin, et~al.]{papamarkou2024position}
Papamarkou, T., Skoularidou, M., Palla, K., Aitchison, L., Arbel, J., Dunson, D., Filippone, M., Fortuin, V., Hennig, P., Hubin, A., et~al.
\newblock Position paper: Bayesian deep learning in the age of large-scale ai.
\newblock \emph{arXiv preprint arXiv:2402.00809}, 2024.

\bibitem[Park et~al.(2023)Park, O'Brien, Cai, Morris, Liang, and Bernstein]{generative_agents}
Park, J.~S., O'Brien, J., Cai, C.~J., Morris, M.~R., Liang, P., and Bernstein, M.~S.
\newblock Generative agents: Interactive simulacra of human behavior.
\newblock In \emph{Proceedings of the 36th Annual ACM Symposium on User Interface Software and Technology}, UIST '23, New York, NY, USA, 2023. Association for Computing Machinery.
\newblock ISBN 9798400701320.
\newblock \doi{10.1145/3586183.3606763}.
\newblock URL \url{https://doi.org/10.1145/3586183.3606763}.

\bibitem[Raffin et~al.(2021)Raffin, Hill, Gleave, Kanervisto, Ernestus, and Dormann]{stable-baselines3}
Raffin, A., Hill, A., Gleave, A., Kanervisto, A., Ernestus, M., and Dormann, N.
\newblock Stable-baselines3: Reliable reinforcement learning implementations.
\newblock \emph{Journal of Machine Learning Research}, 22\penalty0 (268):\penalty0 1--8, 2021.
\newblock URL \url{http://jmlr.org/papers/v22/20-1364.html}.

\bibitem[Reid et~al.(2024)Reid, Savinov, Teplyashin, Lepikhin, Lillicrap, Alayrac, Soricut, Lazaridou, Firat, Schrittwieser, et~al.]{reid2024gemini}
Reid, M., Savinov, N., Teplyashin, D., Lepikhin, D., Lillicrap, T., Alayrac, J.-b., Soricut, R., Lazaridou, A., Firat, O., Schrittwieser, J., et~al.
\newblock Gemini 1.5: Unlocking multimodal understanding across millions of tokens of context.
\newblock \emph{arXiv preprint arXiv:2403.05530}, 2024.

\bibitem[Sasso et~al.(2023)Sasso, Conserva, and Rauber]{sasso2023posterior}
Sasso, R., Conserva, M., and Rauber, P.
\newblock Posterior sampling for deep reinforcement learning.
\newblock In \emph{International Conference on Machine Learning}, pp.\  30042--30061. PMLR, 2023.
\newblock URL \url{https://openreview.net/forum?id=ZwjSECgl6p}.

\bibitem[Schrittwieser et~al.(2020)Schrittwieser, Antonoglou, Hubert, Simonyan, Sifre, Schmitt, Guez, Lockhart, Hassabis, Graepel, et~al.]{Schrittwieser2019-df}
Schrittwieser, J., Antonoglou, I., Hubert, T., Simonyan, K., Sifre, L., Schmitt, S., Guez, A., Lockhart, E., Hassabis, D., Graepel, T., et~al.
\newblock \href{https://www.nature.com/articles/s41586-020-03051-4}{Mastering atari, go, chess and shogi by planning with a learned model}.
\newblock \emph{Nature}, 588\penalty0 (7839):\penalty0 604--609, 2020.

\bibitem[Schulman et~al.(2015)Schulman, Levine, Abbeel, Jordan, and Moritz]{schulman2015trust}
Schulman, J., Levine, S., Abbeel, P., Jordan, M., and Moritz, P.
\newblock Trust region policy optimization.
\newblock In Bach, F. and Blei, D. (eds.), \emph{Proceedings of the 32nd International Conference on Machine Learning}, volume~37 of \emph{Proceedings of Machine Learning Research}, pp.\  1889--1897, Lille, France, 07--09 Jul 2015. PMLR.
\newblock URL \url{https://proceedings.mlr.press/v37/schulman15.html}.

\bibitem[Schulman et~al.(2017)Schulman, Wolski, Dhariwal, Radford, and Klimov]{schulman2017proximal}
Schulman, J., Wolski, F., Dhariwal, P., Radford, A., and Klimov, O.
\newblock Proximal policy optimization algorithms.
\newblock \emph{arXiv preprint arXiv:1707.06347}, 2017.

\bibitem[Sutton \& Barto(2018)Sutton and Barto]{sutton2018reinforcement}
Sutton, R.~S. and Barto, A.~G.
\newblock \emph{Reinforcement learning: An introduction}.
\newblock MIT press, 2018.

\bibitem[Tang et~al.(2024)]{tang2024worldcoder}
Tang, X. et~al.
\newblock Worldcoder: Learning world models by writing code.
\newblock In \emph{Advances in Neural Information Processing Systems (NeurIPS)}, 2024.

\bibitem[Team et~al.(2023)Team, Anil, Borgeaud, Wu, Alayrac, Yu, Soricut, Schalkwyk, Dai, Hauth, et~al.]{team2023gemini}
Team, G., Anil, R., Borgeaud, S., Wu, Y., Alayrac, J.-B., Yu, J., Soricut, R., Schalkwyk, J., Dai, A.~M., Hauth, A., et~al.
\newblock Gemini: a family of highly capable multimodal models.
\newblock \emph{arXiv preprint arXiv:2312.11805}, 2023.

\bibitem[Team et~al.(2024)Team, Mesnard, Hardin, Dadashi, Bhupatiraju, Pathak, Sifre, Rivi{\`e}re, Kale, Love, et~al.]{team2024gemma}
Team, G., Mesnard, T., Hardin, C., Dadashi, R., Bhupatiraju, S., Pathak, S., Sifre, L., Rivi{\`e}re, M., Kale, M.~S., Love, J., et~al.
\newblock Gemma: Open models based on gemini research and technology.
\newblock \emph{arXiv preprint arXiv:2403.08295}, 2024.

\bibitem[Valmeekam et~al.(2023)Valmeekam, Marquez, Sreedharan, and Kambhampati]{valmeekam2023on}
Valmeekam, K., Marquez, M., Sreedharan, S., and Kambhampati, S.
\newblock On the planning abilities of large language models - a critical investigation.
\newblock In \emph{Thirty-seventh Conference on Neural Information Processing Systems}, 2023.
\newblock URL \url{https://openreview.net/forum?id=X6dEqXIsEW}.

\bibitem[Vinyals et~al.(2019)Vinyals, Babuschkin, Czarnecki, Mathieu, Dudzik, Chung, Choi, Powell, Ewalds, Georgiev, et~al.]{vinyals2019grandmaster}
Vinyals, O., Babuschkin, I., Czarnecki, W.~M., Mathieu, M., Dudzik, A., Chung, J., Choi, D.~H., Powell, R., Ewalds, T., Georgiev, P., et~al.
\newblock Grandmaster level in starcraft ii using multi-agent reinforcement learning.
\newblock \emph{Nature}, 575\penalty0 (7782):\penalty0 350--354, 2019.
\newblock URL \url{https://www.nature.com/articles/s41586-019-1724-z}.

\bibitem[Wang et~al.(2024{\natexlab{a}})Wang, Ma, Feng, Zhang, Yang, Zhang, Chen, Tang, Chen, Lin, and et~al.]{llm_survey}
Wang, L., Ma, C., Feng, X., Zhang, Z., Yang, H., Zhang, J., Chen, Z., Tang, J., Chen, X., Lin, Y., and et~al.
\newblock A survey on large language model based autonomous agents.
\newblock \emph{Frontiers of Computer Science}, 18\penalty0 (6), Mar 2024{\natexlab{a}}.
\newblock \doi{10.1007/s11704-024-40231-1}.

\bibitem[Wang et~al.(2024{\natexlab{b}})]{wang2024bytesized32}
Wang, Y. et~al.
\newblock Can language models serve as text-based world simulators?
\newblock \emph{arXiv preprint arXiv:2403.01560}, 2024{\natexlab{b}}.

\bibitem[Yang et~al.(2024)Yang, Du, Ghasemipour, Tompson, Kaelbling, Schuurmans, and Abbeel]{uni_sim}
Yang, S., Du, Y., Ghasemipour, S. K.~S., Tompson, J., Kaelbling, L.~P., Schuurmans, D., and Abbeel, P.
\newblock Learning interactive real-world simulators.
\newblock In \emph{The Twelfth International Conference on Learning Representations}, 2024.
\newblock URL \url{https://openreview.net/forum?id=sFyTZEqmUY}.

\bibitem[Yang et~al.(2023)Yang, Zhou, Li, Tao, Li, Shen, He, Jiang, and Shi]{emma}
Yang, Y., Zhou, T., Li, K., Tao, D., Li, L., Shen, L., He, X., Jiang, J., and Shi, Y.
\newblock Embodied multi-modal agent trained by an llm from a parallel textworld.
\newblock \emph{arXiv preprint arXiv:2311.16714}, 2023.

\bibitem[Yao et~al.(2023{\natexlab{a}})Yao, Yu, Zhao, Shafran, Griffiths, Cao, and Narasimhan]{yao2024tree}
Yao, S., Yu, D., Zhao, J., Shafran, I., Griffiths, T.~L., Cao, Y., and Narasimhan, K.~R.
\newblock Tree of thoughts: Deliberate problem solving with large language models.
\newblock In \emph{Thirty-seventh Conference on Neural Information Processing Systems}, 2023{\natexlab{a}}.
\newblock URL \url{https://openreview.net/forum?id=5Xc1ecxO1h}.

\bibitem[Yao et~al.(2023{\natexlab{b}})]{yao2023tree}
Yao, S. et~al.
\newblock Tree of thoughts: Deliberate problem solving with large language models.
\newblock In \emph{Proceedings of the 40th International Conference on Machine Learning (ICML)}, 2023{\natexlab{b}}.

\end{thebibliography}
